\documentclass[11pt,a4paper]{article}
\usepackage[hyperref]{acl2019}
\aclfinalcopy

\usepackage{tabu} %rowfont

\usepackage{times}
\usepackage{latexsym}
\usepackage{amsfonts}
\usepackage{oplotsymbl}
\usepackage{url}
\usepackage{tabu} %rowfont
\usepackage{bbm}
\usepackage{transparent}
\usepackage[hyperref]{acl2019} 

\usepackage{CJKutf8}
\usepackage{hyperref}
\usepackage{url} 

\usepackage{times}
\usepackage{latexsym}
\usepackage{xcolor}

\usepackage{url}

\usepackage{tikz}
\usepackage{tikz-qtree}
\usepackage{latexsym}
\usepackage{graphicx}			% allows us to import images
\usepackage{times}
\usepackage{latexsym}
\usepackage{mathpartir}
\usepackage{pifont}

\usepackage{proof}
\usepackage{xspace}
\usepackage{multirow}
\usepackage{forest}
\usepackage{subcaption}
\usepackage{diagbox}

\usepackage{dsfont}
\usepackage{amsmath}
\usepackage{amssymb}
\usepackage{amsthm}
\usepackage{algorithm}
\newcommand{\namecite}[1]{\newcite{#1}}
\definecolor{chocolate}{rgb}{0.28, 0.02, 0.03}
\definecolor{PaleGreen}{rgb}{0.33, 0.545,0.33}
\definecolor{colorC0}{RGB}{51,113, 169}
\definecolor{colorC1}{RGB}{243,130,37}
\definecolor{deepmagenta}{rgb}{0.8, 0.0, 0.8}
\definecolor{dollarbill}{rgb}{0.52, 0.73, 0.4}
\definecolor{beaver}{rgb}{0.62, 0.51, 0.44}
\usepackage{array}
\usepackage{hyperref}
\usepackage{url} 

\usepackage{times}
\usepackage{latexsym}
\usepackage{xcolor}

\usepackage{url}

\usepackage{tikz}
\usepackage{tikz-qtree}
\usepackage{latexsym}
\usepackage{graphicx}			% allows us to import images
\usepackage{times}
\usepackage{latexsym}
\usepackage{mathpartir}
\usepackage{pifont}

\usepackage{proof}
\usepackage{xspace}
\usepackage{multirow}
\usepackage{forest}
\usepackage{subcaption}

\usepackage{dsfont}
\usepackage{amsmath}
\usepackage{amssymb}
\usepackage{amsthm}
\usepackage{algorithm}
\usepackage{dblfloatfix}

\newcommand{\state}{(\vecs, \vect)\xspace}
\newcommand{\seq}{(\vecx, \vecy)\xspace}

\newcommand{\delay}{\langle \varepsilon \rangle\xspace}

\newcommand{\pixyabofst}{\pi^\star_{\vecx, \vecy, \alpha, \beta} (\vecs, \vect)\xspace}
\newcommand{\pixyab}{\pi^\star_{\vecx, \vecy, \alpha, \beta}\xspace}

\newcommand{\notes}[1]{}%{\it {\small {#1}}}}

\theoremstyle{definition}

\theoremstyle{plain}

\newcommand{\veca}{\ensuremath{\mathbf{a}}\xspace}

\newcommand{\vecs}{\ensuremath{\mathbf{s}}\xspace}
\newcommand{\vect}{\ensuremath{\mathbf{t}}\xspace}

\newcommand{\ith}[1]{\ensuremath{i^{{th}}}}

\newcount\permx
\newcount\permy
\def\permdot#1#2{
\permx=#1 \advance\permx by-1
\permy=#2 \advance\permy by-1
\psframe[fillcolor=black, fillstyle=solid]
(\permx,\permy)(#1, #2)
}

\newcommand{\boxnum}[1]{{\setlength{\fboxsep}{1pt}\raisebox{1pt}{\hspace{1pt}\fbox{\tiny #1}\hspace{1pt}}}}
\newcommand{\ind}[1]{\ensuremath{_{\kern-0.5pt\boxnum{#1}}}}

\newcommand{\vecx}{\ensuremath{\mathbf{x}}\xspace}
\newcommand{\vecy}{\ensuremath{\mathbf{y}}\xspace}

\def\namecite{\newcite}

\newcommand{\smallnt}[1]{\ensuremath{_{\mbox{\tiny PP}}}\xspace}

\newcommand{\pseudocode}{Algorithm}
\floatname{algorithm}{\pseudocode}

\iffalse

\else

\fi

\title{Simultaneous Translation with Flexible Policy \\
via Restricted Imitation Learning}

\author{Baigong Zheng $^{1,}$\thanks{\; These authors contributed equally.} \,
  Renjie Zheng $^{2, \ast}$ \,
  Mingbo Ma $^{1,\ast}$ \,
  Liang Huang $^{1,2}$
\\
  $^{1}$Baidu Research, Sunnyvale, CA, USA \\
  $^{2}$Oregon State University, Corvallis, OR, USA \\
  {\small \texttt{\{baigongzheng, mingboma\}@baidu.com \qquad zrenj11@gmail.com} } 
}

\date{}

\begin{document}
\begin{CJK}{UTF8}{gbsn}
\maketitle
\begin{abstract}
  Simultaneous translation is widely useful %in many scenarios
  but remains one of the most difficult tasks in NLP.
  Previous work either uses fixed-latency policies, or train
  a complicated two-staged model using reinforcement learning.
  We propose a much simpler {\em single model} that adds a ``delay'' token to the target vocabulary,
  and design a {\em restricted dynamic oracle} to greatly simplify training.
  %is as  easy as local training.
  Experiments on  Chinese$\leftrightarrow$English simultaneous translation
  show that our work leads to flexible policies that achieve better BLEU scores and lower latencies
  compared to
  both fixed and RL-learned policies. % and reinforcement learning method.
\end{abstract}

% !TEX root = main.tex
\section{Introduction}

Simultaneous translation, which translates sentences before they are finished, 
is useful in many scenarios such as international conferences, summits, 
and negotiations.
However, it is widely considered one of the most challenging tasks in NLP,
and one of the holy grails of AI \cite{grissom+:2014}.
A major challenge in simultaneous translation is the word order difference
between the source and target languages, e.g., between SOV languages (German, Japanese, etc.)
and SVO languages (English, Chinese, etc.).

Simultaneous translation is previously studied as a part of real-time speech recognition system~\cite{mahsa+:2013, bangalore+:2012, fugen+:2007, sridhar+:2013, jaitly2016online, graves2013speech}. 
Recently, there have been two encouraging efforts in this problem
with promising but limited success.
\namecite{gu+:2017} propose a complicated two-stage model that is also trained in two stages.
The base model, responsible for producing target words, is a conventional 
full-sentence seq2seq model, 
and on top of that, the READ/WRITE (R/W) model decides, at every step, whether to wait for another source word (READ) or to emit a target word (WRITE)
using the pretrained base model.
This R/W model is trained by reinforcement learning (RL) method without updating the base model.  
\namecite{ma+:2018}, on the other hand, propose a much simpler architecture, which only need one model and can be trained with end-to-end local training method. However, their model follows a fixed-latency policy, which inevitably needs to guess future content during translation.
Table~\ref{tab:expl} gives an example which is difficult for the fixed-latency (wait-$k$) policy but easy for adaptive policy.  
\begin{table}[t]
\resizebox{1\linewidth}{!}{%
\setlength{\tabcolsep}{-2pt}
\centering
\begin{tabu}{ p{1.4cm} |  c c c c c c c c c c c  p{2.5cm} }
\hline 
Chinese &  \ \ 我 \ \ & & 得到 & & 有关 & & 方面 &    & 的 &  & 回应&            \\
\rowfont{\it}
  pinyin & \ \  w\v{o} \ &  & \ \ d\'ed\`ao && y\v{o}ugu\=an && f\=angmi\`an && de & & hu\'iy\`ing &   \\
 gloss & \ \   I \  &  & \ \ receive &  & relevant  && party & & 's  & & response   & \\
 \hline
  wait-1 policy &    & I   & &  received   & & {\color{red}thanks} && {\color{red}from} \ \ & & \ \ relevant  & & parties   \\
  \hline
  wait-5 policy &    &     &        &&  &    &&          & &  I & & received   responses from relevant parties \\
  \hline
 adaptive policy &    & I  &   & received &   &  & &   &  &  &   & responses from relevant parties \\
 \hline
\end{tabu} 
}
\caption{A Chinese-to-English translation example. Wait-1 policy makes a mistake on guessing {\em thanks from} while wait-5 policy has high latency. The adaptive policy can wait for more information to avoid guesses while maintaining low latency.}
\label{tab:expl}
\vspace{-20pt}
\end{table}

We aim to combine the merits of both efforts, that is, 
we design a single model {\em end-to-end trained from scratch} to perform simultaneous translation, as with~\citet{ma+:2018}, which can decide on the fly whether to wait or translate as in~\citet{gu+:2017}.  
There are two key ideas to achieve this:
the first is to add a ``delay'' token (similar to the READ action in \namecite{gu+:2017}, the {\em empty} token in~\citet{press+:2018}, and the `blank' unit in Connectionist Temporal Classification (CTC)~\cite{graves+:2006}) to the target-side vocabulary, and if the model
emits this delay token, it will read one source word;
the second idea is to train the model using (restricted) imitation learning by designing a  (restricted) dynamic oracle as the expert policy.  
Table~\ref{tab:approach} summarizes different approaches for simultaneous translation using neural machine translation (NMT) model.
\begin{table}
  \resizebox{.49\textwidth}{!}{
\begin{tabular}{|p{1.5cm}|p{2.9cm}|p{3cm}|} 
\hline
& seq-to-seq       & prefix-to-prefix   \\ 
\hline
fixed \ \ \ policy    & static Read-Write \; \; \cite{dalvi+:2018} \ \ test-time wait-$k$ \; \; \; \cite{ma+:2018}   &  wait-$k$  \cite{ma+:2018} \\ 
\hline
adaptive \ \ policy & RL \cite{gu+:2017} &  {{\bf imitation learning (this work)}}  \\ 
\hline
\end{tabular}
}
\captionof{table}{Different approaches for simultaneous translation.}
\vspace{-15pt}
\label{tab:approach}
\end{table}

\section{Preliminaries}

Let $\vecx=(x_1, \dots, x_n)$ be a sequence of words.
For an integer $0 \le i \le n$, we denote the sequence consisting of the first consecutive $i-1$ words in $\vecx$ by $\vecx_{< i} = (x_1, \dots, x_{i - 1})$. 
We say such a sequence $\vecx_{< i}$ is a {\em prefix} of the sequence $\vecx$, and define $\vecs \preceq \vecx$ if sequence $\vecs$ is a prefix of $\vecx$. 

\paragraph{Conventional Machine Translation}
Given a sequence $\vecx$ from the source language,
the conventional machine translation model predicts the probability distribution of the next target word $y_j$ at the $j$-th step, conditioned on the full source sequence $\vecx$ and previously generated target words $\vecy_{< j}$,
that is
$p(y_{j} \mid \vecx, \vecy_{< j})$.
The probability of the whole sequence $\vecy$ generated by the model will be 
$p(\vecy \mid \vecx) = \textstyle\prod_{j=1}^{|\vecy|} p(y_j \mid \vecx, \vecy_{< j})$.

To train such a model, we can maximize the probability of ground-truth target sequence conditioned on the corresponding source sequence in a parallel dataset $D$, which is equivalent to minimize the following loss:
\vspace{-5pt}
\begin{equation}\label{equ:full}
\ell(D) = - \textstyle\sum_{(\vecx, \vecy) \in D} \log p(\vecy \mid \vecx).
\vspace{-5pt}
\end{equation}

In this work, we use {\em Transformer}~\cite{vaswani+:2017} as our NMT model, which consists of an encoder and a decoder. The encoder works in a self-attention fashion and maps a sequence of words to a sequence of continuous representations. The decoder performs attention over the predicted words and the output of the encoder to generate next prediction. 
Both encoder and decoder take as input the sum of a word embedding and its corresponding positional embedding.

\paragraph{Prefix-to-Prefix Framework}

Previous work \cite{gu+:2017, dalvi+:2018} use seq2seq models to do simultaneous translation, which are trained with full sentence pairs but need to predict target words based on partial source sentences. 
\namecite{ma+:2018} proposed a {\em prefix-to-prefix} training framework to solve this mismatch. 
The key idea of this framework is to train the model to predict the next target word conditioned on the partial source sequence the model has seen, instead of the full source sequence.

As a simple example in this framework, \namecite{ma+:2018} presented a class of policies, called {\em wait-$k$ policy}, that can be applied with local training in the prefix-to-prefix framework.
For a positive integer $k$, the wait-$k$ policy will wait for the first $k$ source words and then start to alternate generating a target word with receiving a new source word, until there is no more source words, when the problem becomes the same as the full-sequence translation.
The probability of the $j$-th word is $p_k(y_j \mid \vecx_{< j+k}, \vecy_{< j})$, and the probability of the whole predicted sequence is 
$p_k(\vecy \mid \vecx) = \textstyle\prod_{j=1}^{|\vecy|} p_k(y_j \mid \vecx_{< j+k}, \vecy_{< j})$.

% !TEX root = main.tex
\section{Model}

To obtain a flexible and adaptive policy,  we need our model to be able to take both READ and WRITE actions.
Conventional translation model already has the ability to write target words, so we introduce a ``delay'' token $\delay$ in target vocabulary to enable our model to apply the READ action.
Formally, for the target vocabulary $V$, we define an extended vocabulary 
\vspace{-5pt}
\begin{equation}\label{equ:V+}
\vspace{-5pt}
V_{+} = V \cup \{ \delay \}.
\end{equation}
Each word in this set can be an action, which is applied with a transition function $\delta$ on a sequence pair $(\vecs, \vect)$ for a given source sequence $\vecx$ where $\vecs \preceq \vecx$.
We assume $\delay$ cannot be applied with the sequence pair $(\vecs, \vect)$ if $\vecs = \vecx$, then we have the transition function $\delta$ as follows,
\vspace{-7pt}
\begin{multline*}
\delta(\state, a) =  
\left\{
\begin{matrix}
(\vecs \circ x_{|\vecs| + 1}, \vect) & \text{if } a = \delay \\
(\vecs, \vect \circ a) & \text{otherwise} 
\end{matrix}
\vspace{-10pt}
\right. 
\end{multline*}
where $\vecs \circ x$ represents concatenating a sequence $\vecs$ and a word $x$.

Based on this transition function, our model can do simultaneous translation as follows.
Given the currently available source sequence, our model continues predicting next target word until it predicts a delay token. Then it will read a new source word, and continue prediction. Since we use Transformer model, the whole available source sequence needs to be encoded again when reading in a new source word, but the predicted target sequence will  not be changed.

Note that the predicted delay tokens do not provide any semantic information, but may introduce some noise in attention layer during the translation process.
So we propose to remove those delay token in the attention layers except for the current input one.  
However, this removal may reduce the explicit latency information which will affect the predictions of the model since the model cannot observe previous output delay tokens. 
Therefore, to provide this information explicitly, we embed the number of previous delay tokens to a vector and add this to the sum of the word embedding and position embedding as the input of the decoder.

% !TEX root = main.tex
\section{Methods}

\subsection{Training via Restricted Imitation Learning}

We first introduce a restricted dynamic oracle~\cite{cross+huang:2016b} based on our extended vocabulary.
Then we show how to use this dynamic oracle to train a simultaneous translation model via imitation learning.
Note that we do not need to train this oracle.

\paragraph{Restricted Dynamic Oracle}

Given a pair of full sequences $(\vecx, \vecy)$ in data, the input state of our restricted dynamic oracle will be a pair of prefixes $(\vecs, \vect)$ where $\vecs \preceq \vecx$, $\vect \preceq \vecy$ and $(\vecs, \vect) \ne (\vecx, \vecy)$. 
The whole action set is $V_+$ defined in the last section.
The objective of our dynamic oracle is to obtain the full sequence pair $(\vecx, \vecy)$  and maintain a reasonably low latency.  
\begin{figure}%[t]
\centering
\includegraphics[width=7.8cm]{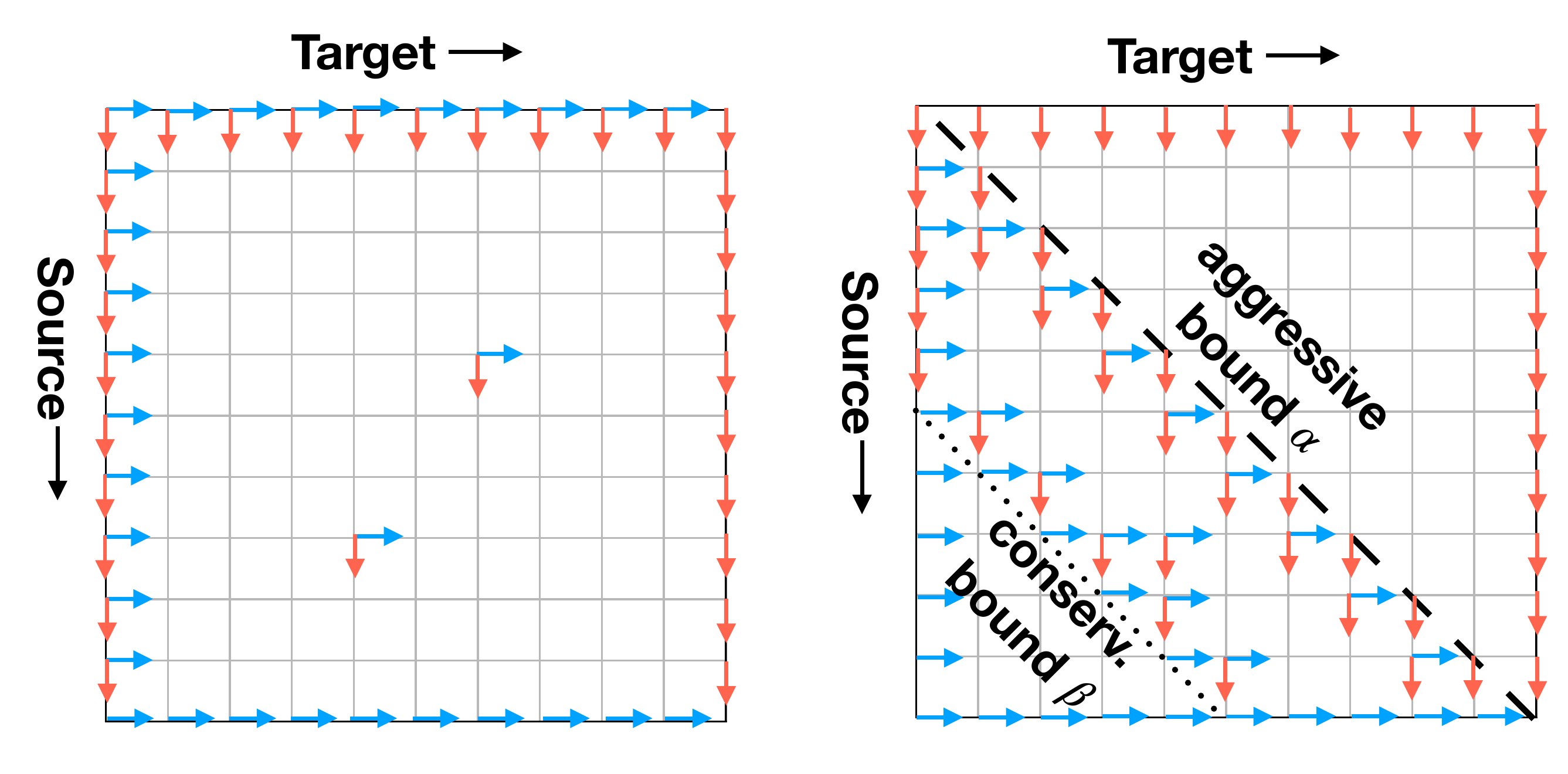}
\captionof{figure}{Illustration of our proposed dynamic oracle on a prefix grid. The blue right arrow represents choosing next ground-truth target word, and the red downward arrow represents choosing the delay token. The left figure shows a simple dynamic oracle without delay constraint. The right figure shows the dynamic oracle with delay constraints.
}
\label{fig:do} 
\vspace{-10pt}
\end{figure}

For a prefix pair $(\vecs, \vect)$, the difference of the lengths of the two prefixes can be used to measure the latency of translation. 
So we would like to bound this difference as a latency constraint.
This idea can be illustrated in the prefix grid~(see Figure~\ref{fig:do}), where we can define a band region and always keep the translation process in this band.
For simplicity, we first assume the two full sequences have the same lengths, i.e. $|\vecx| = |\vecy|$. 
Then we can bound the difference $d = |\vecs| - |\vect|$ by two constants: $\alpha < d < \beta$.
The conservative bound ($\beta$) guarantees relatively small difference and low latency; while the aggressive bound ($\alpha$) guarantees there are not too many target words predicted before seeing enough source words.
Formally, this dynamic oracle is defined as follows.  
\vspace{-12pt}
\begin{multline*}
 \pixyabofst = \\ 
 \left\{
\begin{matrix}
\{ \delay \} & \text{if } \vecs \ne \vecx \text{ and } |\vecs| - |\vect| \le \alpha  \\
\{ y_{|\vect| + 1} \} &  \text{if } \vect \ne \vecy \text{ and } |\vecs| - |\vect| \ge \beta  \\
\{ \delay, y_{|\vect| + 1} \} & \text{otherwise}
\end{matrix}
\right.
\vspace{-14pt}
\end{multline*}

% explain it is optimal
By this definition, we know that this oracle can always find an action sequence to obtain $(\vecx, \vecy)$.
When the input state does not satisfy any latency constraint, then this dynamic oracle will provide only one action, applying which will improve the length difference.
Note that this dynamic oracle is restricted in the sense that it is only defined on the prefix pair instead of any sequence pair. And since we only want to obtain the exact sequence from data, this oracle can only choose the next ground-truth target word other than $\delay$.

% what if |x| != |y|
In many cases, the assumption $|\vecx| = |\vecy|$ does not hold. 
To overcome this limitation, we can utilize the length ratio $\gamma = |\vecx| / |\vecy|$ to modify the length difference: $d' = |\vecs| - \gamma |\vect|$, and use this new difference $d'$ in our dynamic oracle. 
Although we cannot obtain this ratio during testing time, we may use the averaged length ratio obtained from training data~\cite{huang2017finish}.

\paragraph{Training with Restricted Dynamic Oracle}

We apply imitation learning to train our translation model, using the proposed dynamic oracle as the expert policy.
Recall that the prediction of our model depends on the whole generated prefix including $\delay$ (as the input contains the embedding of the number of $\delay$), which is also an action sequence. 
If an action sequence $\veca$ is obtained from our oracle, then applying this sequence will result in a prefix pair, say $\vecs_{\veca}$ and $\vect_{\veca}$, of $\vecx$ and $\vecy$.
Let $p(a \mid \vecs_{\veca}, \vect_{\veca})$ be the probability of choosing action $a$ given the prefix pair obtained by applying action sequence $\veca$.
Then the averaged probability of choosing the oracle actions conditioned on the action sequence $\veca$ will be \\ 
\begin{equation*} 
f(\veca , \pixyab) = \frac{\textstyle \sum\limits_{a \in \pixyab(\vecs_{\veca}, \vect_{\veca})} p(a \mid \vecs_{\veca}, \vect_{\veca})}{|\pixyab(\vecs_{\veca}, \vect_{\veca})|}.
\end{equation*}

To train a model to learn from the dynamic oracle, we can sample from our oracle to obtain a set, say $S(\vecx, \vecy)$, of action sequences for a sentence pair $(\vecx, \vecy)$. 
The loss function for each sampled sequence $\veca \in S(\vecx, \vecy)$ will be
\vspace{-5pt}
\begin{equation*}
\vspace{-5pt}
\ell(\veca | \vecx, \vecy) = -\textstyle\sum\limits_{i=1}^{|\veca|} \log f(\veca_{<i} , \pixyab).
\end{equation*}
For a parallel text $D$, the training loss is 
\vspace{-3pt}
\begin{equation*}
\vspace{-5pt}
\ell(D) = \textstyle\sum\limits_{(\vecx, \vecy) \in D} \textstyle\sum\limits_{\veca \in S(\vecx, \vecy)}\frac{1}{|S(\vecx, \vecy)|} \ell(\veca | \vecx, \vecy).
\end{equation*}

Directly optimizing the above loss may require too much computation resource since for each pair of $\seq$, the size of $S(\vecx, \vecy)$ (i.e. the number of different action sequences) can be exponentially large.
To reduce the computation cost, we propose to use two special action sequences as our sample set so that our model can learn to do translation within the two latency constraints.  
Recall that the latency constraints of our dynamic oracle $\pixyab$ are defined by two bounds: $\alpha$ and $\beta$. For each bound, there is a unique action sequence, which corresponds to a path in the prefix grid, such that following it can generate the most number of prefix pairs that make this bound tight.  
Let $\veca^{\alpha}_{\seq}$ ($\veca^{\beta}_{\seq}$) be such an action sequence for $\seq$ and $\alpha$ ($\beta$).
We replace $S(\vecx, \vecy)$ with $\{ \veca^{\alpha}_{\seq}, \veca^{\beta}_{\seq} \}$, then the above loss for dataset $D$ becomes
\vspace{-3pt}
\begin{equation*} 
\vspace{-3pt}
\ell_{\alpha, \beta}(D) = \textstyle\sum\limits_{(\vecx, \vecy) \in D} \frac{\ell(\veca^{\alpha}_{\seq} | \vecx, \vecy) + \ell(\veca^{\beta}_{\seq} | \vecx, \vecy)}{2}.
\end{equation*}
This is the loss we use in our training process.

Note that there are some steps where our oracle will return two actions, so for such steps we will have a multi-label classification problem where labels are the actions from our oracle.
In such cases, Sigmoid function for each action is more appropriate than the Softmax function for the actions will not compete each other~\cite{Ma2017GroupSC, zheng2018multi, ma2019learning}.
Therefore, we apply Sigmoid for each action instead of using Softmax function to generate a distribution for all actions.

% !TEX root = main.tex

\subsection{Decoding}

We observed that the model trained on the two special action sequences occasionally violates the latency constraints and visits states outside of the designated band in prefix grid. To avoid such case, we force the model to choose actions such that it will always satisfy the latency constraints. That is, if the model reaches the aggressive bound, it must choose a target word other than $\delay$ with highest score, even if $\delay$ has higher score; if the model reaches the conservative bound, it can only choose $\delay$ at that step.
We also apply a temperature constant $e^t$ to the score of $\delay$, which can implicitly control the latency of our model without retraining it. This improves the flexibility of our trained model so that it can be used in different scenarios with different latency requirements.

% !TEX root = main.tex

\section{Experiments}

\begin{figure*}
\centering
\begin{minipage}[b]{.45 \linewidth}
\begin{subfigure}[b]{.9\textwidth}
\centering
\includegraphics[height=4.2cm]{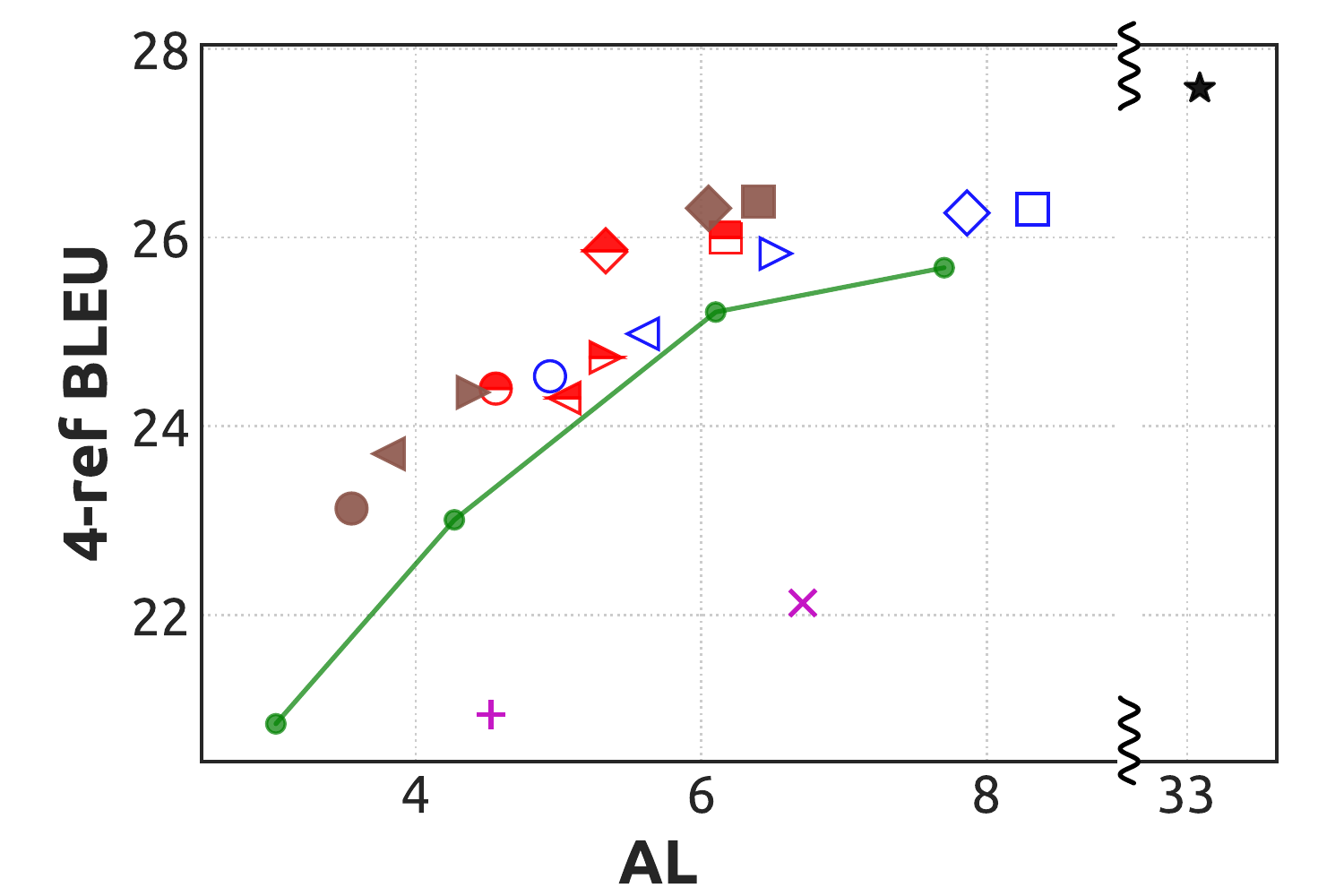}
\caption{\label{fig:a} Chinese-to-English}
\end{subfigure}
\end{minipage}\qquad
\begin{minipage}[b]{.45 \linewidth}
\begin{subfigure}[b]{.9\textwidth}
\centering
\includegraphics[height=4.2cm]{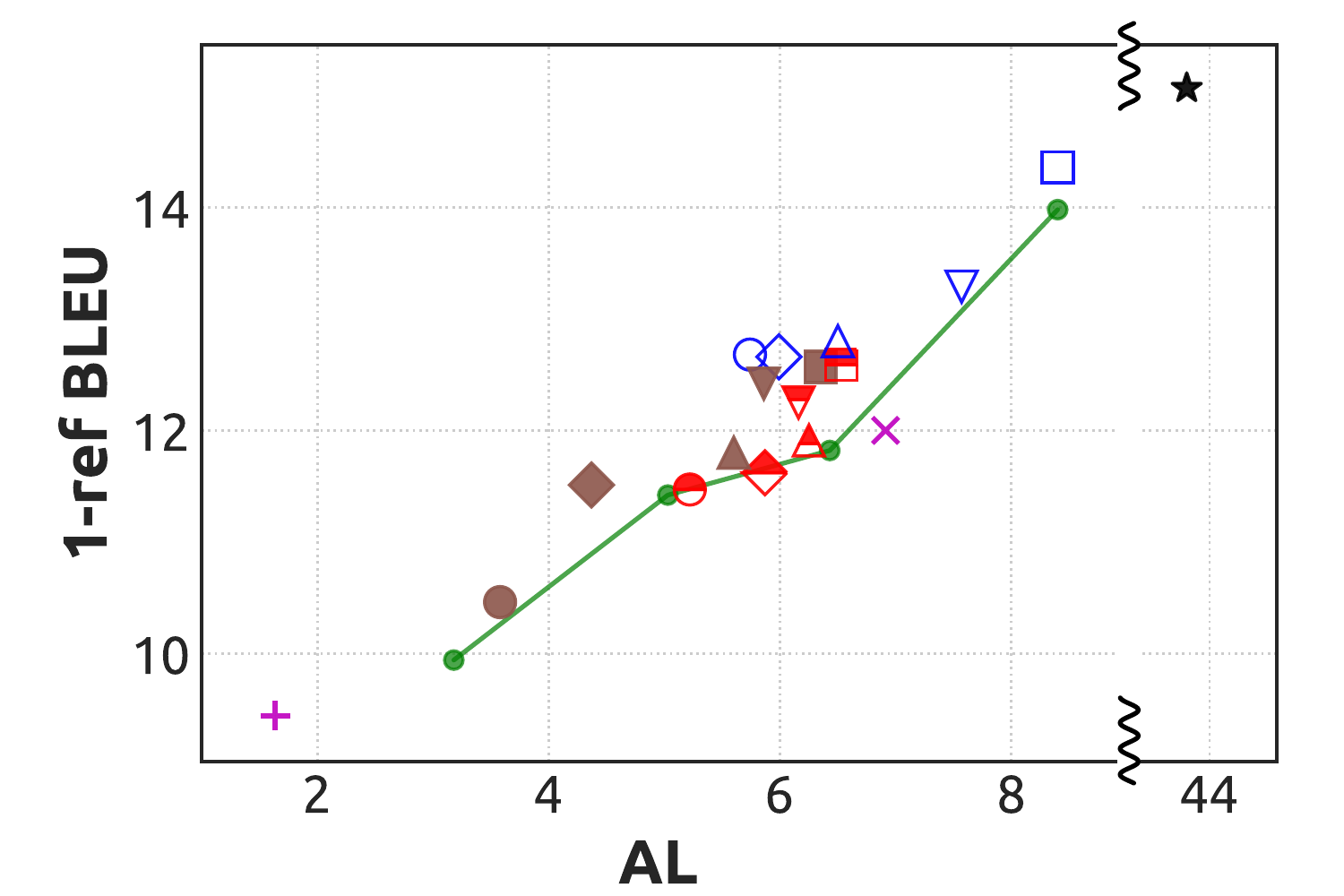}
\caption{\label{fig:b} English-to-Chinese}
\end{subfigure}
\end{minipage} 
\caption{
\begin{minipage}[t]{.40\linewidth}
\vspace{-20pt}
\small
Translation quality (BLEU) against latency (AL) on testing sets.
Markers \textcolor{dollarbill}{$\circletfill$}: wait-$k$ models for $k \in \{1, 3, 5, 7\}$,
\textcolor{deepmagenta}{$+$}: RL with CW = 5,
\textcolor{deepmagenta}{$\times$}: RL with CW = 8,
\textcolor{black}{$\bigstar$}: full-sentence translation.
Markers for our models are given in the right table.
\end{minipage} \quad
\begin{minipage}[t]{.45\linewidth}
\resizebox{1.0\textwidth}{!}{%
\begin{tabular}{|c|c |c |c |c |c |c|c|c|}
\hline
\multicolumn{2}{|c|}{Training} & \multicolumn{7}{c|}{Decoding Policy} \\
\hline
$\alpha$ & $\beta$ & wait-$\alpha$ & wait-$\beta$ & $t=-2$ &$t=-0.5$ &$t=0$ & $t = 4.5$ & $t=9$ \\
\hline 
1 & 5 & \textcolor{beaver}{$\circletfill$} & \textcolor{beaver}{$\squadfill$} & \textcolor{beaver}{$\triangleplfill$} & \textcolor{beaver}{$\triangleprfill$} & \textcolor{beaver}{$\rhombusfill$} & \textcolor{beaver}{$\trianglepafill$} & \textcolor{beaver}{$\trianglepbfill$} \\
\hline 
3 & 5 & \textcolor{red}{$\circletfillha$} & \textcolor{red}{$\squadfillha$} & \textcolor{red}{$\triangleplfillhr$} & \textcolor{red}{$\triangleprfillhl$} & \textcolor{red}{$\rhombusfillha$} & \textcolor{red}{$\trianglepafillha$} & \textcolor{red}{$\trianglepbfillhb$} \\
\hline 
3 & 7 & \textcolor{blue}{$\circlet$} & \textcolor{blue}{$\squad$} & \textcolor{blue}{$\trianglepl$} & \textcolor{blue}{$\trianglepr$} & \textcolor{blue}{$\rhombus$} & \textcolor{blue}{$\trianglepa$} & \textcolor{blue}{$\trianglepb$} \\
\hline 
\end{tabular} 
}
\end{minipage}
}
\label{fig:exp}
\vspace{-10pt}
\end{figure*}

\begin{table*}[!b]
\resizebox{0.99\linewidth}{!}{%
\setlength{\tabcolsep}{0pt}
\centering
\begin{tabu}{ c |  c c c c c c c c c c c c c c c    l }
\hline 
Chinese & 一 & 名 &  不 &   & 愿 & & 具名 & & 的 && 欧盟 &    & 官员 &          & 指出 &    \ \ \   ... \\
\rowfont{\it}
  pinyin &  y\`i&m\'ing & b\'u &  & y\`uan && j\`um\'ing && de && \=Oum\'eng & & g\=uany\'uan &  &zh\v{\i}ch\=u&    \\
 gloss &   \; a &-& not &  & willing && named &  & 's && EU & & official  &&point out&  \ \ \   ...\\
 \hline
  wait-3  &    & &   & a   &   &   {\color{red}us}   & &\!official \!&& who\!& &\!\!declined\!\! & & \!\!to\!\! & & \ \ be     named    said  that   ... \\
  \hline
 our work\; &    & &   & a   &   &  &    & &  &   &  & {\color{blue}eu} &   & official &     & \ \ ,    who declined  to be   named  ,   pointed out ... \\
 \hline
\end{tabu} 
}
\caption{A Chinese-to-English development set example. Our model is trained with $\alpha=3$ and $\beta=7$.}
\label{tb:olympic}
\end{table*}

To investigate the empirical performance of our proposed method,
we conduct experiments on NIST corpus for Chinese-English.
We use NIST 06 (616 sentence pairs) as our development set and NIST 08 (691 sentence pairs) as our testing set.  
We apply tokenization and byte-pair encoding (BPE)~\cite{sennrich+:2015} on both source and target languages to reduce their vocabularies.
For training data, we only include 1 million sentence pairs with length larger than 50.
We use Transformer~\cite{vaswani+:2017} as our NMT model, and   
our implementation is adapted from PyTorch-based OpenNMT \cite{klein+:2017}.
The architecture of our Transformer model is the same as the base model in the original paper.

We use BLEU \cite{BLEU:2002} as the translation quality metric and {\em Average Lagging} (AL) introduced by~\citet{ma+:2018} as our latency metrics,
which measures the average delayed words.
AL avoids some limitations of other existing metrics, such as
insensitivity to actual lagging like Consecutive Wait (CW)~\cite{gu+:2017},
and sensitivity to input length like Average Proportion (AP)~\cite{Cho+:16} .

\paragraph{Results} 
We tried three different pairs for $\alpha$ and $\beta$: (1, 5), (3, 5) and (3, 7), and summarize the results on testing sets in Figure~\ref{fig:exp}.  
Figure~\ref{fig:exp} (a) shows the results on Chinese-to-English translation.  
In this direction, our model can always achieve higher BLEU scores with the same latency, compared with the wait-$k$ models and RL models.
We notice the model prefers conservative policy during decoding time when $t=0$.
So we apply negative values of $t$ to encourage the model to choose actions other than $\delay$. This can effectively reduce latency without sacrificing much translation quality, implying that our model can implicitly control latency during testing time.

Figure~\ref{fig:exp} (b) shows our results on English-to-Chinese translation.
Since the English source sentences are always longer than the Chinese sentences, we utilize the length ratio $\gamma = 1.25$ (derived from the dev set) during training, which is the same as using ``catchup'' with frequency $c=0.25$ introduced by~\namecite{ma+:2018}.  
Different from the other direction, models for this direction works better if the difference of $\alpha$ and $\beta$ is bigger. 
Another difference is that our model prefers aggressive policy instead of conservative policy when $t=0$.
Thus, we apply positive values of $t$ to encourage it to choose $\delay$, obtaining more conservative policies to improve translation quality.

\paragraph{Example}

We provide an example from the development set of Chinese-to-English translation in Table~\ref{tb:olympic} to compare the behaviours of different models.  
Our model is trained with $\alpha=3, \beta=7$ and tested with $t=0$.
It shows that our model can wait for information ``{\it \=Oum\'eng}'' to translates ``eu'', 
while the wait-3 model is forced to guess this information and made a mistake on the wrong guess ``us'' before seeing ``{\it \=Oum\'eng}''.

\paragraph{Ablation Study}
To analyze the effects of proposed techniques on the performance,
we also provide an ablation study on those techniques for our model trained with $\alpha=3$ and $\beta=5$ in Chinese-to-English translation.  
The results are given in Table~\ref{tab:ablation}, and
show that all the techniques are important to the final performance and using Sigmoid function is critical to learn adaptive policy.
\begin{table}[h]\centering
\resizebox{.49\textwidth}{!}{%
\begin{tabular}{|p{3cm}|c|c|c|c|c|c|}\hline
\multirow{3}{*}{Model}          & \multicolumn{6}{c|}{Decoding Policy} \\ \cline{2-7}
& \multicolumn{2}{c|}{Wait-$3$ } & \multicolumn{2}{c|}{Wait-$5$ } & \multicolumn{2}{c|}{t=0} \\ \cline{2-7}
                    & BLEU  & AL   & BLEU     & AL     & BLEU  & AL       \\\hline
          Wait-$3$  & 29.32 & 4.60 & -        & -      & -     & -        \\\hline
          Wait-$5$  &   -   &  -   & 30.97    & 6.30   & -     & -        \\\hline
 keep \; $\delay$ \;  in \; \; \; \;  attention  & 29.55 & 4.50 & 30.68    & 6.49   & 30.74 & 6.53     \\\hline
 no $\delay$ number \; \; embedding      & 30.20 & 4.76 & 30.98    & 6.36   & 30.65 & 6.29     \\\hline
 use Softmax \; \; \ \ \ \ \ instead of Sigmoid         & 29.23 & 5.11 & 31.46    & 6.79   & 29.99 & 4.79     \\\hline
 Full               & 29.45 & 4.71 & 31.72    & 6.35   & 31.59 & 6.28     \\\hline
\end{tabular}
}
\caption{Ablation study on Chinese-to-English development set with $\alpha=3$ and  $\beta=5$.  }
\label{tab:ablation} 
\vspace{-10pt}
\end{table}

\vspace{-5pt}
\section{Conclusions}
We have presented a simple model that includes a delay token in the target vocabulary such that the model can apply both READ and WRITE actions during translation process without a explicit policy model.
We also designed a restricted dynamic oracle for the simultaneous translation problem and provided a local training method utilizing this dynamic oracle.
The model trained with this method can learn a flexible policy for simultaneous translation and achieve better translation quality and lower latency compared to previous methods.

\newpage

\bibliographystyle{acl_natbib}
\bibliography{main}

\end{CJK}

\end{document}